\documentclass[aoas2]{imsart}

\usepackage{graphicx} 
\usepackage{subfigure}

\usepackage{natbib}

\usepackage{algorithm}
\usepackage{algorithmic}
\usepackage{hyperref}

\usepackage{amsmath}
\usepackage{amssymb}
\usepackage{algorithm}
\usepackage{algorithmic}
\usepackage{bbm}
\usepackage{color}
\usepackage{subfigure}
\usepackage{hyperref}
\usepackage{url}
\usepackage{natbib}

\newcommand{\rom}[1]{\uppercase\expandafter{\romannumeral #1\relax}}

\begin{document}

\begin{frontmatter}

\title{Lock in Feedback in Sequential Experiments}
\runtitle{Lock in Feedback}

\begin{aug}
\author{\fnms{Maurits} \snm{Kaptein}}
and
\author{\fnms{Davide} \snm{Iannuzzi}\thanksref{t1}}
\affiliation{Radboud University, Nijmegen, the Netherlands,\\and Vrije Universiteit, Amsterdam, the Netherlands}
\runauthor{Kaptein \& Iannuzzi}
\end{aug}

\begin{abstract}
We often encounter situations in which an experimenter wants to find, by sequential experimentation, $x_{max} = \arg\max_{x} f(x)$, where $f(x)$ is a (possibly unknown) function of a well controllable variable $x$. Taking inspiration from physics and engineering, we have designed a new method to address this problem. In this paper, we first introduce the method in continuous time, and then present two algorithms for use in sequential experiments. Through a series of simulation studies, we show that the method is effective for finding maxima of unknown functions by experimentation, even when the maximum of the functions drifts or when the signal to noise ratio is low.
\end{abstract}

\thankstext{t1}{D.I. acknowledges the support of the European Research Council (grant agreement n. 615170) and of the Stichting voor Fundamenteel Onderzoek der Materie (FOM), which is financially supported by the Netherlands organization for scientific research (NWO).}

\end{frontmatter}

\section{Introduction}

\begin{figure}[!ht]
 \centering
    \includegraphics[width=0.85\textwidth]{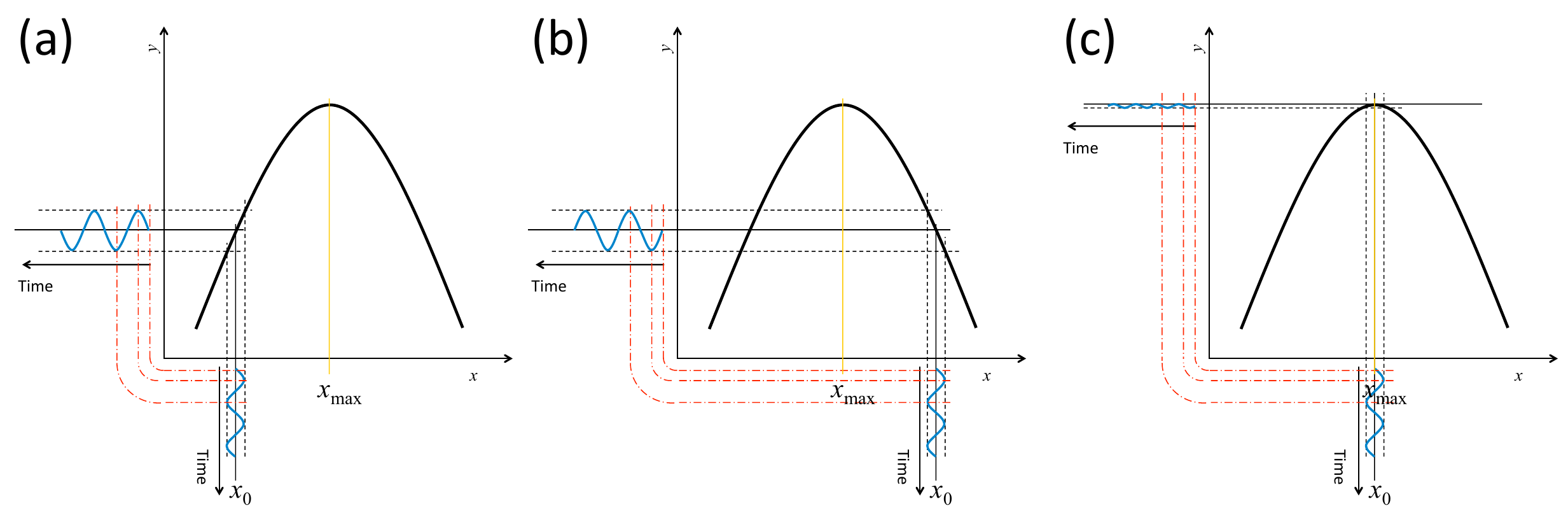}
           \caption{Illustration of the lock-in principle used in physics and engineering to maintain a bring and maintain an independent, controllable variable $x$ onto the value $x_{max}$ for which a dependent variable $y$ is maximized. The value of $x$ is oscillated sinusoidally around a central value $x_0$. (a): If $x_{0}<x_{max}$, $y$ oscillates at the same frequency as $x$, in phase (i.e., a maximum value of $x$ corresponds to a maximum value of $y$). (b): If $x_0>x_{max}$, $y$ oscillates again at the same frequency as $x$, but with opposite phase (i.e., a maximum value of $x$ corresponds to a minimum value of $y$). (c): If $x_0=x_{max}$, $y$ ceases to oscillate at the frequency of $x$, but starts to oscillate at a doubled frequency. Lock-in amplifiers can detect the amplitude and the phase of the oscillation at a reference frequency, and, therefore, indicate whether $x$ is smaller, larger, or equal to $x_0$.}
    \label{fig:intro}
\end{figure}

When designing an experiment where a given parameter must be kept constant throughout the entire duration of the measurement, physicists and engineers often rely on feedback techniques that, in real time, can properly re-adjust the configuration of the experiment to compensate for unexpected drifts  \citep{Scofield1994}. Fig. \ref{fig:intro} illustrates, for instance, a well-established approach that is used to maintain a variable $x$ always locked at the value that maximizes the value of another variable $y$, which is some function -- possibly with large noise -- of $x$. The algorithm behind this approach, which will be described more in depth later in the text, is based on the following steps:

\begin{itemize}
\item[1] Fix a central value $x_0$ of the variable $x$;
\item[2] Add an oscillation of amplitude $A$ at a fixed angular frequency $\omega$: $x=x_0+A_x\cos\left( \omega t \right)$.
\item[3] Measure the amplitude of the oscillations that the variable $y$ has, in response of the oscillation of the variable $x$, at the same angular frequency $\omega$, and further measure whether the oscillation are in phase or out of phase;
\item[4] Set a new value of $x_0$, adding (if the oscillation of $y$ are in phase with the oscillation of $x$) or subtracting (if the oscillation of $y$ are out of phase with respect to the oscillation of $x$) a value proportional to the value measured in step 4: $x_{0,new}=x_0\pm \gamma A_y$, where $\gamma$ is a constant. Iterate steps 2 to 4 for the whole duration of the experiment.
\end{itemize}

The above described feedback loop pushes the value of $x_0$ closer and closer to the value $x_{max}$ that maximizes $y$. As $x_0$ approaches $x_{max}$, the oscillations in $y$ become smaller and smaller, moving $x_0$ in a series of steps of decreasing size. Finally, when $x_0=x_{max}$, the variable $y$ ceases to oscillate at frequency $\omega$: because of this $x_0$ can stay locked in on $x_{max}$. However, if the curve suddenly shifts to another position (e.g., if the relationship between $x$ and $y$ changes, a phenomenon referred to as concept drift \citep{Gaber2005, Anagnostopoulos2012a}), the $\omega$ component of $y$ becomes different from zero again, forcing the feedback loop to move the value of $x_0$ towards the new value of $x_{max}$. Hence, the feedback loop enables one to hold on to the value of $x$ sequentially that maximizes the value of $y$.

Interestingly, the feedback loop described above can work well even if the variable $y$ is affected by a high degree of noise. To extract the signal at frequency $\omega$, in fact, one can make use of a commercial instrument called \textit{lock-in amplifier}, which rejects all the components of the signals that do not beat at the frequency of interest.  The algorithm used by a lock-in amplifier can of course be applied to digital (discrete timepoints) data as well. It is thus worth asking whether the approach adopted in a lock-in amplifier may be used in other contexts where, in the presence of a highly noisy set of data, one wants to maintain one variable locked to the value that maximizes the value of another.

Tantalized by this opportunity, we propose here to use lock-in feedback (LiF) algorithms for the optimization of the price in (e.g.,) a rebate action. The idea is to present each customer a different price, which is changed sinusoidally around a central value, causing the revenue to oscillate at the same frequency. As the customers take their purchasing decision, a lock-in algorithm monitors the oscillations of the revenue at the price oscillation frequency. Like in the feedback loop described above, the central value of the price is continuously adjusted until the revenue ceases to oscillate at that frequency. At this point, in fact, the revenue is maximum (price elasticity = 1). If an unexpected event moves the price elasticity curve, the algorithm will automatically push the central price towards the new maximizing value.

Next to product pricing of rebate actions, many more examples could be conceived in the social sciences:
\begin{itemize}
\item In economics, firms might be able to manipulate the price $x$ of an offering and subsequently observe their revenue $y$. Here a firm seeks to find the value of $x$ that maximizes $y$ \citep[for examples see][]{Kung2002, Jiang2011}.
\item In industry, the outcome $y$ of a business process might depend on the amount of some raw material $x$ used in the process.
\item In communication research, a communication professional might seek to find the length of an email message $x$ that leads to the highest number of clicks $y$ on a link in that message \citep{Ansari2005}.
\item In medicine, a physician seeks to find the optimal dose $x$ of a medicine to maximize the health outcome $y$ of her patients \citep[see, e.g.,][]{Sapareto1984, Marschner2007}.
\item In education, scholars might seek to select learning tasks which are quantified by their difficulty $x$, that have the highest effect on learning $y$ of their pupils.
\end{itemize}
In the above cases the functional form of $f(x)$ is often not known, the outcome $y$ is observed with noise, and likely the treatment values that maximize the outcome are subject to concept drift \citep{Gaber2005, Anagnostopoulos2012a} (thus, they change over time). Here we present a method to find $x_{max}$ which does not require an explicit specification of $f(x)$ or its derivatives, performs well in the face of noise, and is robust to concept drift. 

To prove the merits of LiF in such cases, we have performed an extensive numerical exercise that simulates the performance of LiF in a diverse range of situations, including ones where the observed signal is merely the choice of a consumer to yes or no adopt a product for a given (rebate) price; a scenario directly in line with the pricing challenges as identified above. We show that, in the presence of the noise induced by the variance of the willingness to pay across the population of the customers entering the shop, our lock-in algorithm allows the seller to both determine and maintain the price that optimizes the revenue of the shop. Furthermore, we demonstrate that if the price elasticity curve changes, the algorithm can detect the direction of the change and converge again to the optimal price. 

It has to be noted that it is a well-known and well-studied challenge to find optimal (according to some specified criterion) treatment values in (sequential) experiments. This challenge is acknowledged in many branches of science and engineering \citep[see, e.g.,][]{allen2003experimental, bardsley1996optimal, kuck2006smc}. An often researched topic is that of design optimization (DO), in which experimental designs are identified that lead to the smallest possible variances in the estimated model parameters \citep{burnetas1996optimal,McClelland1997}. More recently, an interest in adaptive design optimization (ADO) methods \citep{myung2009optimal,myung2013tutorial} and sequential experimentation methods has emerged: researchers are looking for effective ways to sequentially determine optimal treatment values in experiments as the experimental data is being collected \citep{zhang2010optimal}. Notably, work on Multi Armed Bandit (MAB) problems \citep[e.g.,][]{Lai1987, Whittle1980, Scott2010, Bubeck2011, Yue2012} and stochastic optimization \citep[e.g.,][]{NIPS2011_4475} has led to efficient sequential sampling schemes for various experimental designs and optimization criteria. 

This paper however introduces a novel sequential sampling scheme for a specific sequential design problem: we examine the problem in which the treatment values are continuous (e.g., with $x$ being $\in \mathbb{R}$) and the researcher seeks a treatment value $x_{max}$ at which the observed outcome $y$---which, at least in part, depends on $x$---obtains its maximal value. Thus we examine the situation in which an experimenter wants to find, by sequential experimentation, $x_{max} = \arg\max_{x} f(x)$, where $f(x)$ is a (possibly unknown) function of a well controllable variable $x$ and is likely observed with noise. We focus on the simple case where $x$ is a scalar. In the remainder of the paper, we index sequential trials by $t \in \{1, \dots, \mathcal{T} \}$. Our ultimate aim is to describe an experimental method for manipulating $x_t$ (in discrete time) to find, sequentially, the value of $x$ that maximizes $y$.

The current manuscript is structured as follows: first, we briefly review the literature on DO, MAB problems, and stochastic optimization to position our method. Next, we discuss LiF as a solution to the treatment optimization problem considered in this paper. LiF is based on a solution that is routinely implemented in physics and engineering applications which relies on the idea of systematically changing the value of the treatment in time via so-called \textit{lock-in amplifier techniques} \citep{Scofield1994}. We introduce its basic principles in continuous time. Subsequently, we present two algorithms to use LiF in sequential experiments. We then, by simulation, compare the two algorithms, and examine the performance of LiF in several scenario's of signal-to-noise ratio and in situations of concept drift. Furthermore, we examine the use of LiF in cases in which the observable outcome is discrete; which is for example the case in the optimization of prices as described above. Finally, we examine the empirical \emph{regret} -- the search cost of the algorithm compared to an algorithm which has full information -- of the proposed procedure and compare it to a standard solution in the MAB literature \citep{berry1985bandit}. 

\subsection{Treatment optimization methods}

The problem of finding $x_{max}$ is treated in a number of branches in the experimental design and machine learning literature. The problem can be approached as an optimal design problem, in which the main aim is to design an experiment that efficiently provides us with information regarding $f(\vec{x})$ \citep[see, e.g.,][]{o2003gentle, Myung2009}. Often, in the DO literature, experiments are treated statically, and the functional form of the data generating function is assumed known: the remaining question is to determine the optimal treatments given a fixed size of the experiment and an assumed relationship to precisely estimate the parameters of interest. 

Recently, \citep{myung2013tutorial} introduced an advanced method of DO into the psychology literature called Adaptive Design Optimization (ADO). The aim of ADO is to create adaptive experiments which are optimized to distinguish between competing explanations of the data \citep{myung2009optimal}. However, in this literature the main aim is to find treatment values to efficiently estimate parameters given a number of model assumptions. Instead, our focus is on efficiently finding treatment values which maximize some observable outcome of the experiment.

Sequentially finding optimal treatments, where optimal is defined in terms of observed outcomes, is explicitly studied in the MAB literature \citep{berry1985bandit}. In this problem specification researchers consider policies $\mathcal{P}$ which describe how to select actions $a \in \mathcal{A}$ (the treatment values) at different times $t$ where the aim is to maximize the cumulative reward $R(t) =  \sum_{t=1}^T r_i$  \citep{Bubeck2011}. The reward is assumed to be a function, possibly with noise, of the actions. Many specifications of the MAB problem exists: researchers have considered independent treatments (the traditional $k$-armed bandit problem \citep{Whittle1980}), related treatments, continuous treatments, etc. \citep{Audibert2009, Bubeck2011b}. The MAB problem, and its generalization, the contextual MAB problem \citep{Li2010a, Beygelzimer2011} present an active area of research in the machine learning literature.

The literature on stochastic optimization with bandit feedback \citep{NIPS2011_4475, agarwal2010optimal} considers the problem of finding the optimal value of continuous treatments \citep{flaxman2005online}. Of special interest for the current proposal are derivative-free (or gradient-free) methods in which the gradient of the function (which is of use for e.g., (stochastic) gradient descent method) is assumed unknown and is itself approximated during the sequential experiment \citep{shamir2012complexity}. In this paper we present a derivative free method to perform stochastic optimization with bandit feedback. The presented method is well-suited for practical use in sequential experiments due to its ease of implementation: in the current paper we provide several algorithms for performing the optimization in real-life settings. Before presenting our novel sequential approach to solving the continuous treatment optimization problem, we first introduce its theoretical background assuming that the treatment does not vary in discrete sequential steps, but rather can be varied continuously (in continuous time).

\section{Finding the maximum of a curve with a lock-in algorithm}

In this section we detail the basic principles behind LiF assuming continuous time in which $x$ can be manipulated. Let's assume that $y$ is a continuous function $f$ of $x$: $y=f(x)$. Let's further assume that $x$ oscillates with time according to:
\begin{align}
\label{oscillation}
x(t)	&=x_0+A\cos\left( \omega t \right)
\end{align}
where $\omega$ is the angular frequency of the oscillation, $x_0$ its central value, and $A$ its amplitude. For relatively small values of $A$, Taylor expanding $f(x)$ around $x_0$ to the second order, one obtains:
\begin{align}
\label{taylor}
\begin{split}
y(x(t)) &=f(x_0)+\left( x_0+A \cos\left( \omega t \right) - x_0 \right) \left( \left. \frac{\partial f}{\partial x}\right|_{x=x_0} \right) \\
	  & +\frac{1}{2}\left( x_0+A \cos\left( \omega t \right) - x_0 \right)^2 \left( \left. \frac{\partial^2 f}{\partial x^2}\right|_{x=x_0} \right)
\end{split}
\end{align}
which can be simplified to:
\begin{align}
\label{evidence}
\begin{split}
y(x(t))	&=k+A \cos\left(\omega t\right)\left( \left. \frac{\partial f}{\partial x}\right|_{x=x_0}\right) \\
		& +\frac{1}{4}A ^2\cos\left(2\omega t\right)\left(\left.  \frac{\partial^2 f}{\partial x^2}\right|_{x=x_0}\right)
\end{split}
\end{align}
where $k=f(x_0)+1/4 A^2 \left(\left. \partial^2 f / \partial x^2 \right|_{x=x_0} \right)$. It is thus evident that, for small oscillations, $y$ becomes the sum of three terms: a constant term, a term oscillating at angular frequency $\omega$, and a term oscillating at angular frequency $2\omega$.

Suppose we ourselves can actively manipulate $x$ and measure $y$, and that $f$ is continuous and only has one maximum and no minimum.\footnote{For simplicity of exposure we only consider these well-behaved functions in this paper.} Further suppose that one is interested to find the value $\arg\max_{x} y = f(x)$ which we denote with $x_{max}$, and that our measurements of $y$ contain noise
\begin{align}
\label{real}
y(t)	&= f(x(t))+ \epsilon_t
\end{align}
where $\epsilon$ denotes the noise and $\epsilon \sim \pi()$ where $\pi$ is some probability density function and $\mathbb{E}[\epsilon | x] = 0$.

Following the scheme used in physical lock-in amplifiers \citep[see, e.g.,][]{Scofield1994}, we multiply the observed $y$ variable by $\cos\left( \omega t \right)$. Using eq. \ref{evidence} and eq. \ref{real}, one obtains:
\begin{align}
\begin{split}
\label{complete}
y_\omega(t) &= \cos\left(\omega t \right) \biggl[ k+A \cos\left(\omega t\right)\left( \left. \frac{\partial f}{\partial x}\right|_{x=x_0}\right)  \\
		    & +\frac{1}{4}A ^2\cos\left(2\omega t\right)\left(\left.  \frac{\partial^2 f}{\partial x^2}\right|_{x=x_0}\right) + \epsilon \biggr]
\end{split}
\end{align}
where $y_\omega$ is the value of $y$ after it has been multiplied by  $\cos \left( \omega t \right)$. Eq. \ref{complete} can be written more compactly as:
\begin{align}
\label{rewrite}
\begin{split}
y_\omega	& = \frac{A}{2}\left( \left. \frac{\partial f}{\partial x}\right|_{x=x_0} \right) \\
		& + k_\omega \cos\left( \omega t\right) + k_{2\omega}\cos\left( 2\omega t \right) \\
		& +k_{3\omega}\cos\left(3\omega t \right)+ \epsilon \cos\left(\omega t \right)
\end{split}
\end{align}
where
\begin{align}
k_\omega 		& = k+A^2/8\left(\left. \partial^2 f / \partial x^2 \right|_{x=x_0} \right) \\
k_{2\omega}	& =A/2\left(\left. \partial^2 f / \partial x^2 \right|_{x=x_0} \right)  \\
 k_{3\omega}	& =A^2/8\left(\left. \partial^2 f / \partial x^2 \right|_{x=x_0} \right).
\end{align}
Integrating $y_\omega$ over a time $T=\frac{2\pi N}{\omega}$, where $N$ is a positive integer and $T$ denotes the time needed to integrate $N$ full oscillations, one obtains:
\begin{align}
\label{rewrite2}
y_\omega^{*}=\frac{TA}{2}\left( \left. \frac{\partial f}{\partial x}\right|_{x=x_0} \right)+\int_0^T \epsilon \cos\left(\omega t \right) dt
\end{align}

Depending on the noise level, one can tailor the integration time, $T$, in such a way to reduce the second addendum of the right hand of eq. \ref{rewrite2} to negligible levels, effectively averaging out the noise in the measurements. Under those circumstances, $y_\omega^{*}$ provides a direct measurement of the value of the first derivative of $f$ at $x=x_0$.

The above method thus yields quantitative information regarding the first derivative of $f$ at $x=x_0$, providing, in this way, a logical update strategy of $x_0$: if $y_\omega^{*}<0$, then $x_0$ is larger than the value of $x$ that maximizes $f$; likewise, if $y_\omega^{*}>0$, $x_0$ is smaller than the value of $x$ that maximizes $f$. Thus, based on the oscillation observed in $y_\omega$ we are now able to move $x_0$ closer to $x = \arg\max_{x} f(x)$ using an update rule $x_0 := x_0 + \gamma y_\omega^{*}$ where $\gamma$ quantifies the learn rate of the procedure. Hence, we can setup a feedback loop that allows us to keep $x_0$ close to $x_{max}$, even if $f(x)$ changes over time.

Note that, multiplying $y$ by $\cos{2\omega t}$ and using a similar approach as the one described above to extract the amplitude of the oscillation of $y$ at frequency $2\omega$, one would be able to measure the second derivative of the function $f$ at $x=x_0$. This property can be useful when, for instance, $f(x)$ is known to be an exact parabola to not only derive the direction of the step towards the maximum, but to work out the exact step size (see Appendix \ref{app:exact}).

\section{Algorithm for LiF in discrete time}

In practical terms, measurements can never run in continuous mode. Therefore, we now present an algorithm for LiF in discrete time. To simplify notation, we will index sequential measurements by $y_t$ where $t=1, \dots, t=\mathcal{T}$ where $\mathcal{T}$ denotes the length---possibly infinite---of the experiment that is ran to find $\arg\max_{x} f(x)$.

In discrete time we can use the same procedure as above in which we start with $x_0$, and for each sample oscillate around $x_0$ with a known frequency $\omega$ and known amplitude $A$:
\begin{align}
x_t = x_0 + A \cos{\omega t}
\end{align}
which will result in measurements given by
\begin{align}
y_t = f(x_0 + A \cos{\omega t}) + \epsilon_t
\end{align}

On the basis of the arguments reported above, we can now implement a feedback loop that iteratively adjusts the value of $x_0$ until $x$ reaches $x_{max}$. After that, if the function $f$ changes, the loop can follow the value of $x$ to the new maximizing position and thus stay ``locked''. The procedure is similar to that given in Equation \ref{rewrite} and \ref{rewrite2}, where we first multiply the outcome $y_t$ by $\cos(\omega t)$ and subsequently integrate out the noise term (summing in the discrete case). In the following sections we present two possible implementations for LiF in discrete time for use in sequential experiments.

\subsection{LiF-\rom{1}: Batch updates of $x_0$}

Our first implementation of LiF (denoted LiF-\rom{1}) is presented in Algorithm \ref{Alg:LiF1}. In this implementation we summate observations $y_t$, which we multiply by $\cos(\omega t)$, for a batch period of length $T$, after which we update $x_0$. Variable $y_{\omega}^{\Sigma}$ contains a running sum of $y_t \cos{\omega t}$ over $t$ that is used for the integration.

\begin{algorithm}
\caption{Implementation of LiF-\rom{1} for single variable maximization in data stream using a batch approach.}
\label{Alg:LiF1}
\begin{algorithmic}
\REQUIRE $x_0$, $A$, $T$, $\gamma$, $y_{\omega}^{\Sigma} = 0$				
\STATE $\omega = \frac{2 \pi}{T}$											
\FOR{$t=1, \dots, \mathcal{T}$}
	\STATE $x_t = x_0 + A \cos{\omega t}$									
	\STATE $y_t = f(x_0 + A \cos{\omega t}) + \epsilon_t$						
	\STATE $y_{\omega}^{\Sigma} = y_{\omega}^{\Sigma} + y_t \cos{\omega t}$		
	\IF{$(t \mod T == 0)$}
		\STATE $y_{\omega}^{*} =  y_{\omega}^{\Sigma} / T$					
		\STATE $x_0 = x_0 + \gamma y_{\omega}^{*} 	$						
		\STATE $y_{\omega}^{\Sigma} = 0$									
	\ENDIF
\ENDFOR
\end{algorithmic}
\end{algorithm}

The tuning parameters for LiF-\rom{1}, which should be set by the experimenter, are  $x_0$, $A$, $T$, $\gamma$. Here below we describe some general criteria the choice may be based on:
\begin{itemize}
\item It is advised to set $x_0$ as close as possible to $x_{max}$. The choice can only be based on the available information on $f$. The more accurate the information, the closer the initial $x_0$ to $x_{max}$, the faster the convergence of the loop to $x_{max}$.
\item The amplitude $A$ affects the costs of the search procedure, because a large $A$ implies querying a large range of $x$ values with (possibly) low resulting $y$ values. However, $A$ also influence the learning speed: a very small $A$ leads to small updates steps, while a large value of $A$ might lead to a value of $\gamma y_{\omega}^{*}$ that ``overshoots'' $x_{max}$.
\item The integration time $T$ affects the variability of the update of $x_0$, with larger integration times leading to a smoother update but slower convergence.
\item The learn-rate $\gamma < 1$ determines the step size at each update of $x_0$. This can be interpreted, and tuned, akin learn-rates in, for instance, stochastic gradient descent methods \citep{Poggio2011}.
\end{itemize}

\subsection{LiF-\rom{2}: Continuous updates of $x_0$}

For some applications the batch updates of $x_0$ -- as implied by the continuous time analysis and defined in Algorithm \ref{Alg:LiF1} -- might not be feasible. Algorithm \ref{Alg:LiF2} presents a modified version of LiF (denoted LiF-\rom{2}) in which $x_0$ is updated every observation. LiF-\rom{2} starts by filling up a buffer of length $T$ which we denote by the vector $\vec{y}_{\omega} = \{NA_1, \dots, NA_T\}$, after which each observation leads to an update of $x_0$. In the algorithm description the values $y_{t-T}, \dots, y_t$ are stored in the vector $\vec{y}_{\omega}$. By defining the learn rate as $\frac{\gamma}{T}$ the tuning parameters in LiF-\rom{2} are the same as those discussed for LiF-\rom{1}.

\begin{algorithm}
\caption{Implementation of LiF-\rom{2} for single variable maximization using continuous updates.}
\label{Alg:LiF2}
\begin{algorithmic}
\REQUIRE $x_0$, $A$, $T$, $\gamma$, $\vec{y}_{\omega} = \{NA_1, \dots, NA_T\}$	
\STATE $\omega = \frac{2 \pi}{T}$											
\FOR{$t=1, \dots, \mathcal{T}$}
	\STATE $x_t = x_0 + A \cos{\omega t}$									
	\STATE $y_t = f(x_0 + A \cos{\omega t}) + \epsilon_t$						
	\STATE $\vec{y}_{\omega}  = \texttt{push}(\vec{y}_{\omega} , y_t \cos{\omega t})$	
	\IF{$(t > T)$}
		\STATE $y_{\omega}^{*} =  (\sum \vec{y}_{\omega}) / T$					
		\STATE $x_0 = x_0 + \frac{\gamma}{T} y_{\omega}^{*} 	$				
	\ENDIF
\ENDFOR
\end{algorithmic}
\end{algorithm}

\section{Simulation study 1: Comparison of Batched and streaming LiF and examination of tuning parameters}
\label{sec:1sims}

In this section we study, by simulation, the differences between LiF-\rom{1} and LiF-\rom{2}, and the effects of the tuning parameters  $A$, $T$, and $\gamma$ in a situation in which $y=f(x)$ is measured without noise.

\begin{figure}[!ht]
 \centering
   \includegraphics[width=0.75\textwidth]{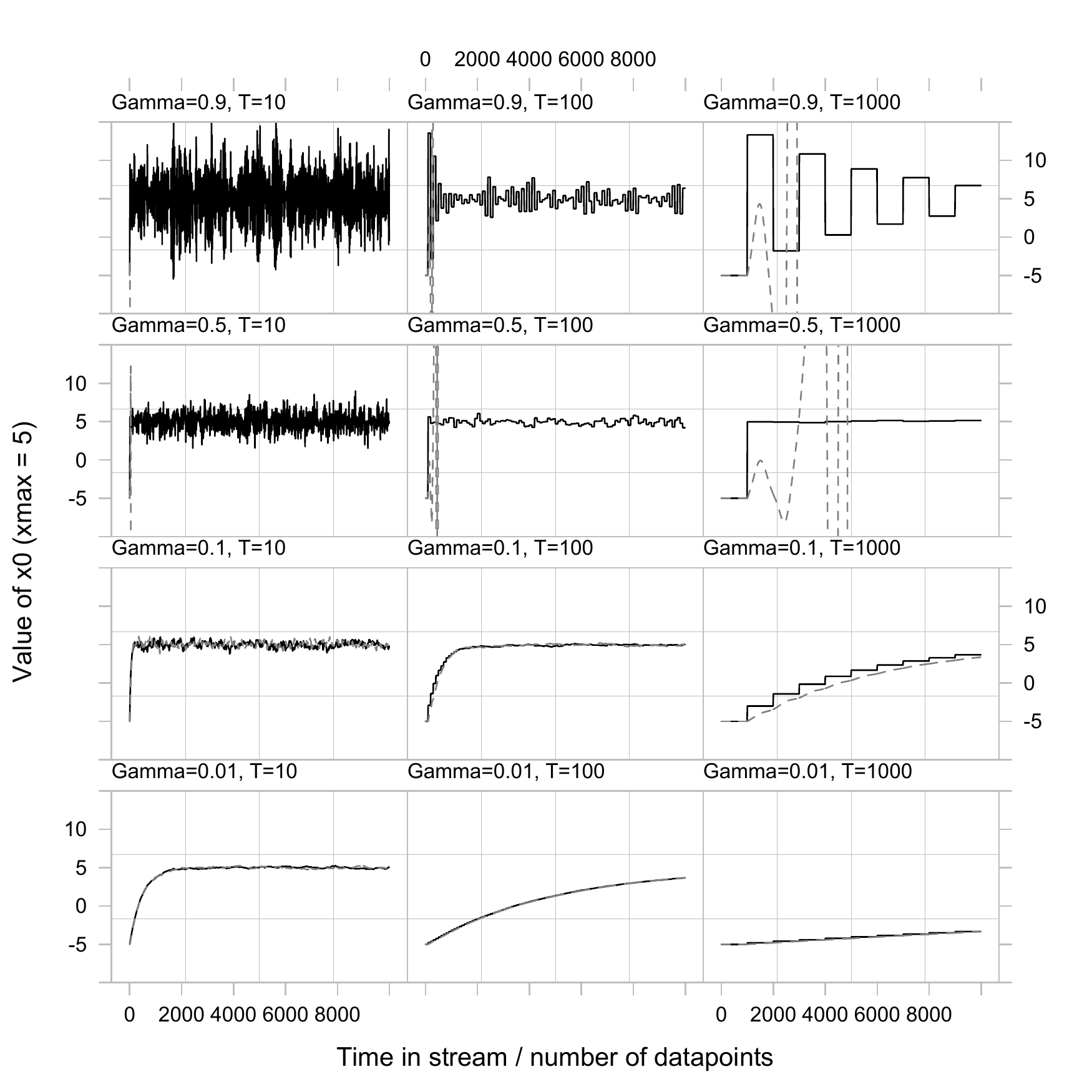}
	 \caption{Examination of the effect of LiF tuning parameters $\gamma$ and $T$ for $A$=1. Displayed are the results for LiF-\rom{1} (black solid line) and LiF-\rom{2} (gray dotted line)}
       \label{fig1}
\end{figure}

\begin{figure}[!ht]
 \centering
    \includegraphics[width=0.75\textwidth]{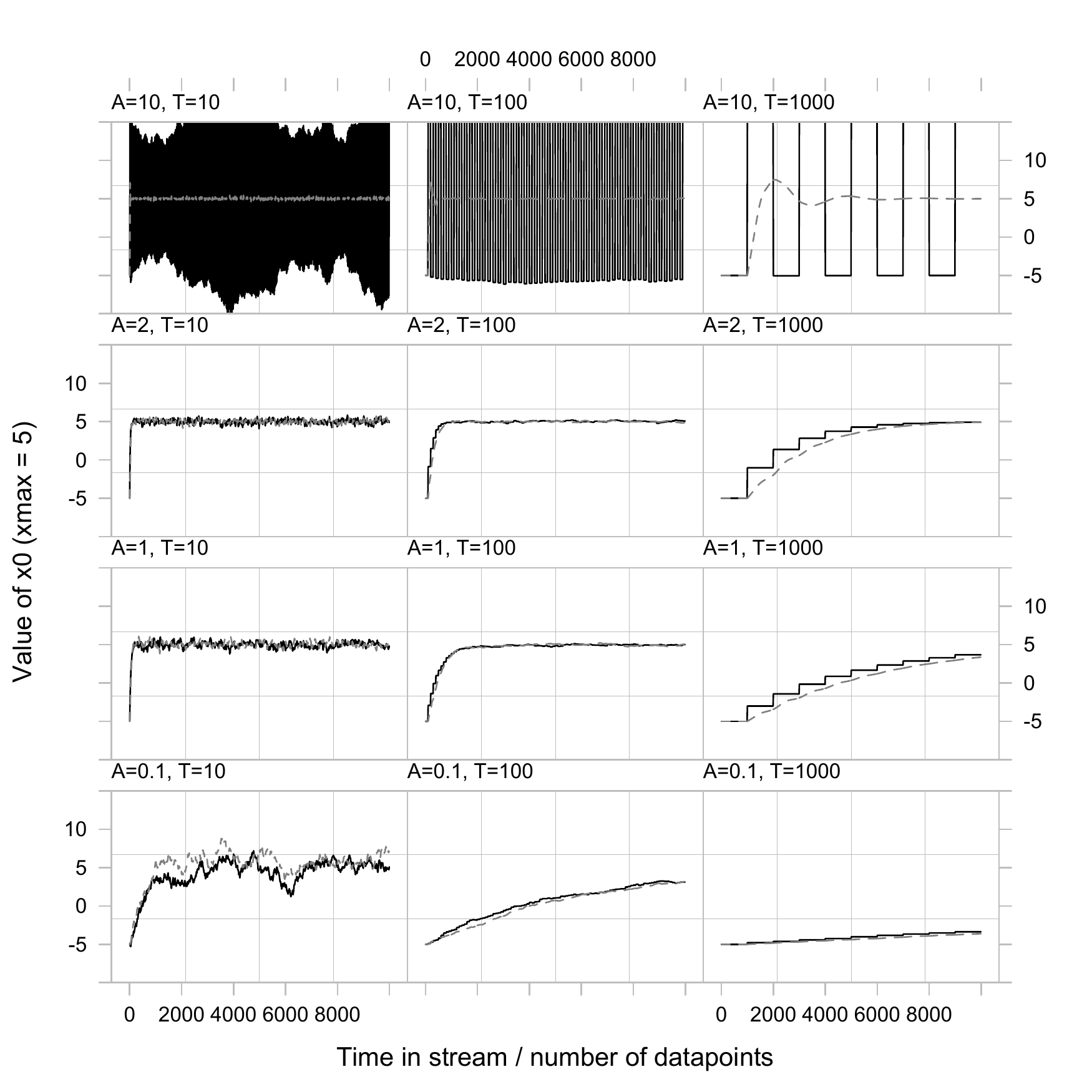}
           \caption{Examination of the effect of tuning parameters $A$ and $T$ for $\gamma=.1$. Displayed are the results for LiF-\rom{1} (black solid line) and LiF-\rom{2} (gray dotted line)}
    \label{fig2}
\end{figure}

Figure \ref{fig1} presents the performance of both LiF-\rom{1} and LiF-\rom{2} for data generated using
\begin{align}
\label{generate:model}
f(x) & = -2(x-5)^2 + \epsilon
\end{align}
where $\epsilon \sim \mathcal{N}(0,0)$ and obviously $x_{max} = 5$. The figure displays the performance of LiF for $\mathcal{T}=10000$ using the following tuning parameter settings
\begin{itemize}
\item $x_0 = -5$.
\item $T \in \{10, 100, 100\}$
\item $A = 1$ 
\item $\gamma \in \{.01, .1, .5, .9\}$
\end{itemize}
The rows of Figure \ref{fig1} (top to bottom) present decreasing values of $\gamma$, while the columns (left to right) present increasing values of $T$. We fix $A=1$. Each panel presents the value of $x_0$ during the data stream as selected using LiF-\rom{1} (black solid line) and LiF-\rom{2} (gray dotted line). It is clear that LiF can ``overshoot'' the maximum for values of $\gamma$ that are too high (top two rows). This happens for both LiF-\rom{1} and LiF-\rom{2}, although LiF-\rom{1} seems more robust. For small values of $\gamma$ the performance of the algorithms is very similar, and increases in the integration window $T$ merely smooth the updating procedure.

In Figure \ref{fig2} the results are plotted for the same setup, but this time we vary $A \in \{.1, 1, 2, 10\}$, while we fix $\gamma = .1$. Here it is clear that for large values of $A$ LiF-\rom{1} has a tendency to become unstable (see top rows), while the streaming LiF-\rom{2} is much more robust for erroneous selection of $A$. Very small choices for the amplitude $A$ lead to very slow updates of $x_0$ in both cases. Again, increased in $T$ merely smooth the process. The simulations give an impression of the importance of the tuning parameters $x_0$, $A$, $T$, $\gamma$ and their relationships. In the remainder of this paper we will focus on the evaluation -- through simulation -- of the performance of LiF-\rom{2} in cases of noise and concept drift.

\section{Simulation study 2: Effects of noise}

\begin{figure}[!ht]
 \centering
    \includegraphics[width=0.75\textwidth]{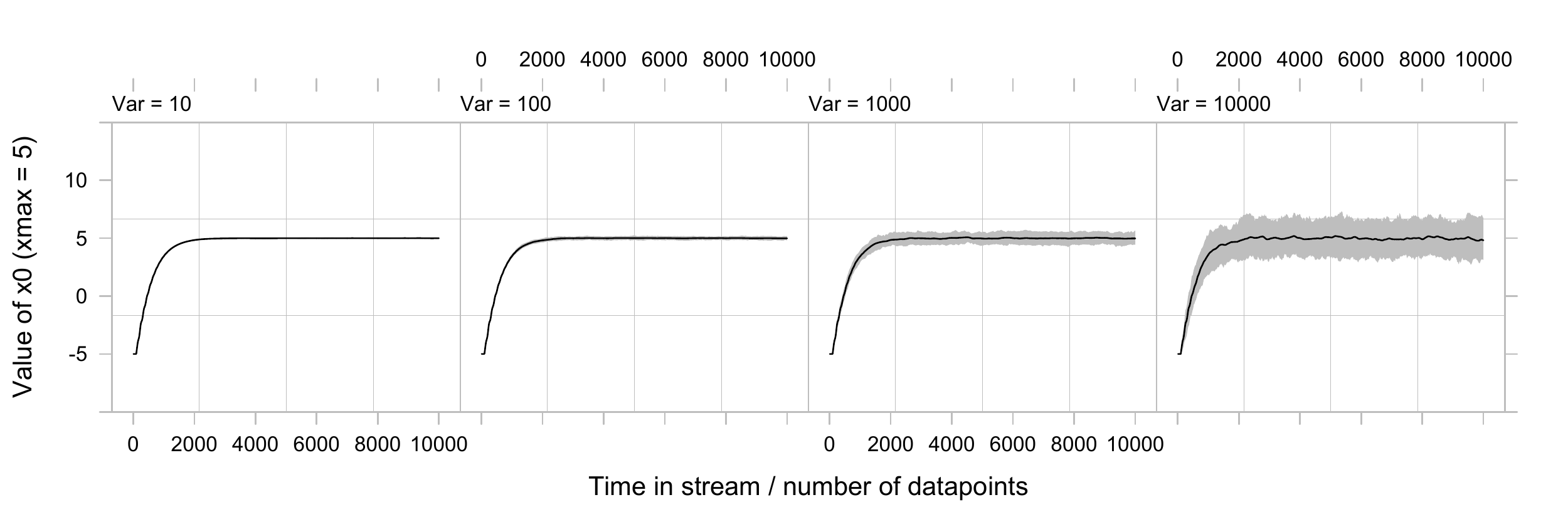}
      \caption{Examination of the effect of different levels of noise $\sigma^2 \in \{10,100,1000, 10000\}$. Note that LiF performs very well also in the presence of noise (see text for more details).}
          \label{fig3}
\end{figure}

To examine the impact of (measurement) noise on the performance of LiF-\rom{2} we repeat the simulations as described in Simulation Study 1 using the data generating model described by Equation \ref{generate:model} with $\epsilon \sim \mathcal{N}(0,\sigma^2)$ and $\sigma^2 \in \{10,100,1000,10000\}$. We choose tuning parameters: $x_0=-5$, $A=1$, $T=100$, $\gamma=.1$. Contrary to the simulations presented in Section \ref{sec:1sims} we now repeat the procedure $m=100$ times: Figure \ref{fig3} presents the average $x_0$ over the $100$ simulation runs as well as the $95\%$ confidence bounds. From Figure \ref{fig3} it is clear that LiF-\rom{2} performs very well in the face of noise.

\section{Simulation study 3: Performance of LiF-\rom{2} in cases of concept drift}

One of the advantages of Lock in Feedback as opposed to other methods of finding $x_{max}$ is the fact that LiF can also be used to find a maximum of a function in cases of concept drift \citep{Gaber2005}: even when $f(x)$ changes over time, LiF provides a method to keep the value of the treatment $x$ close to $x_{max}$.

To illustrate this latter advantage of LiF-\rom{2} we setup a simulation using the following data generating model:
\begin{align}
\label{generate:model2}
f(x,t) & = -2( (x-.0025t) - 5)^2 + \epsilon
\end{align}
where the $(x-.0025t)$ term ensures that during the stream running from $t=0$ to $t=10^4=\mathcal{T}$ the value of $x_{max}$ moves from $5$ to $30$. We choose $x_0=-20$ (note the different starting position compared to the previous simulations), $A=1$, $T=100$, $\gamma=.1$ and $\sigma^2 = 10$. We investigate the performance of LiF-\rom{2} in this case of concept drift.

\begin{figure}[!ht]
  \centering
    \includegraphics[width=0.65\textwidth]{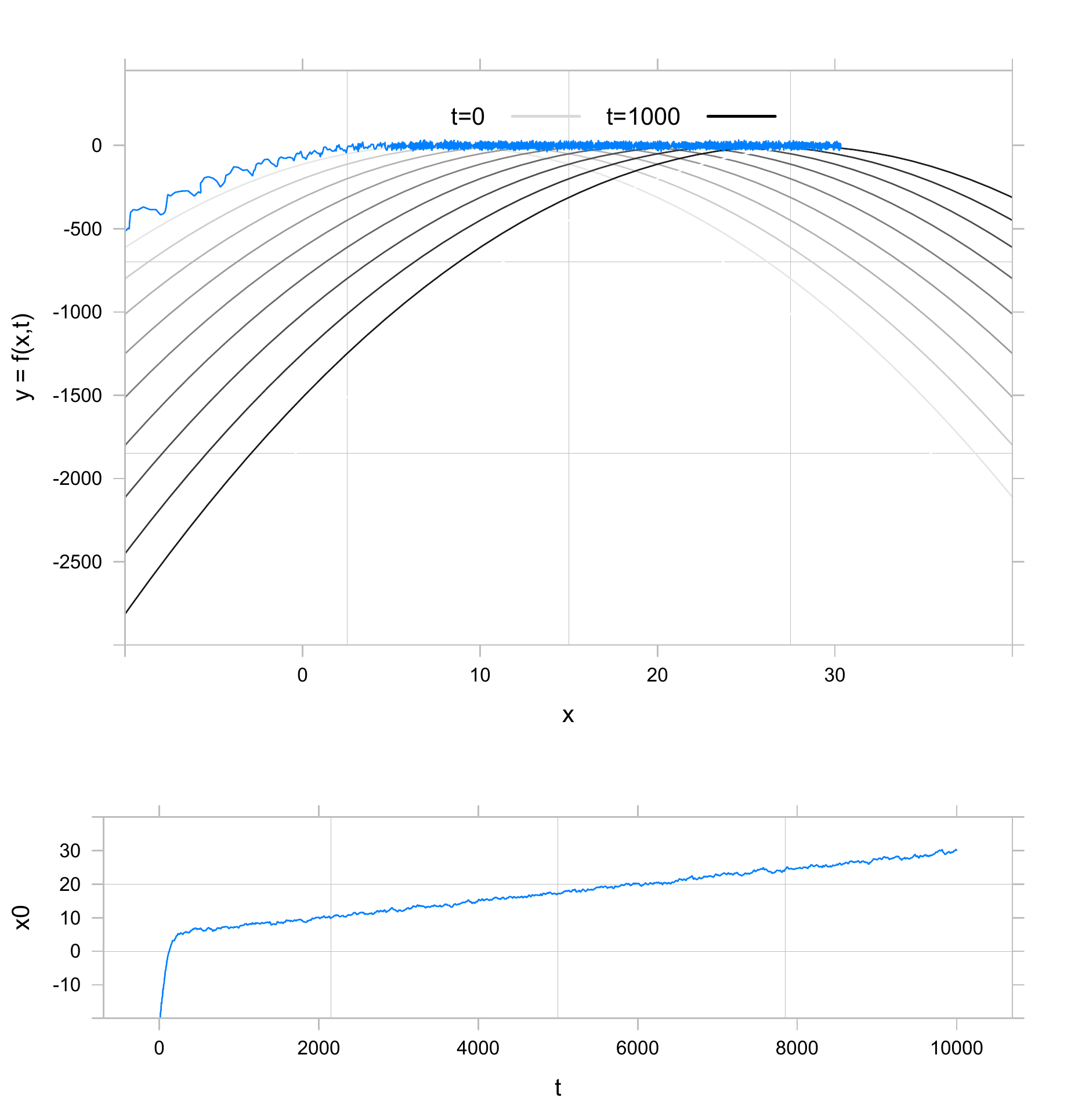}
      \caption{Illustration of LiF in the case of concept drift. As the true maximum shifts (top panel) LiF is able to follow the maximum and keep $x_0$ close to $x_{max}$ (bottom panel).}
          \label{fig:concept}
\end{figure}

Figure \ref{fig:concept} presents in the top panel $y=f(x,t)$ for distinct values of $t \in \{0, 1000, \dots, 10000\}$ in different shades of grey. The concept drift is illustrated by the different locations of the parabola. Superimposed in blue is the value of $x_0$ as selected by LiF-\rom{2}. In the bottom panel the value of $x_0$ as a function of the length of the stream is presented. It is clear that LiF-\rom{2} quickly finds $x_{max}$ and follows the maximum as it moves during the stream.

\section{Simulation study 4: Dichotomous observations}

In the introduction we described as a use case of our proposed method the optimization of sales prices to maximize the revenue. This specific case presents a novel problem since the dependent variable $y$, encoding the purchase decision of a customer after a price has been pitched is dichotomous, and the actual outcome of interest---if the firm aims to maximize its revenue---is a function of the observable and the manipulated variable $r(t)= \sum_{i=1}^{t} y_i x_i$. Since $y_i \in \{0,1\}$, the signal $r(t)$ used as an optimization criteria contains a different type of noise; while the expected value $Pr(y=1 | x) \times x$ of an offer could be approximated, the data itself contains non-zero values only when the decision is made to purchase a product.

To empirically examine the performance of LiF in such a setting we setup a simulation study in which we assume that the data generating model looks as follows:
\begin{align}
\label{eq:logit}
y_t & \sim \texttt{Bernoulli}\left(\frac{1}{1 + e^{-(10-x_t)}}\right)\\
r_t & = y_t x_t
\end{align}
Intuitively the above specification indicates that the probability that a consumer chooses to buy a product decreases as the price, $x$, of the product increases, while the (expected) revenue is computed using the probability of a purchase given a specific price multiplied by that price. Given this setup the (expected) $x_{max}$ is approximately $8$. 

\begin{figure}[!ht]
  \centering
    \includegraphics[width=0.8\textwidth]{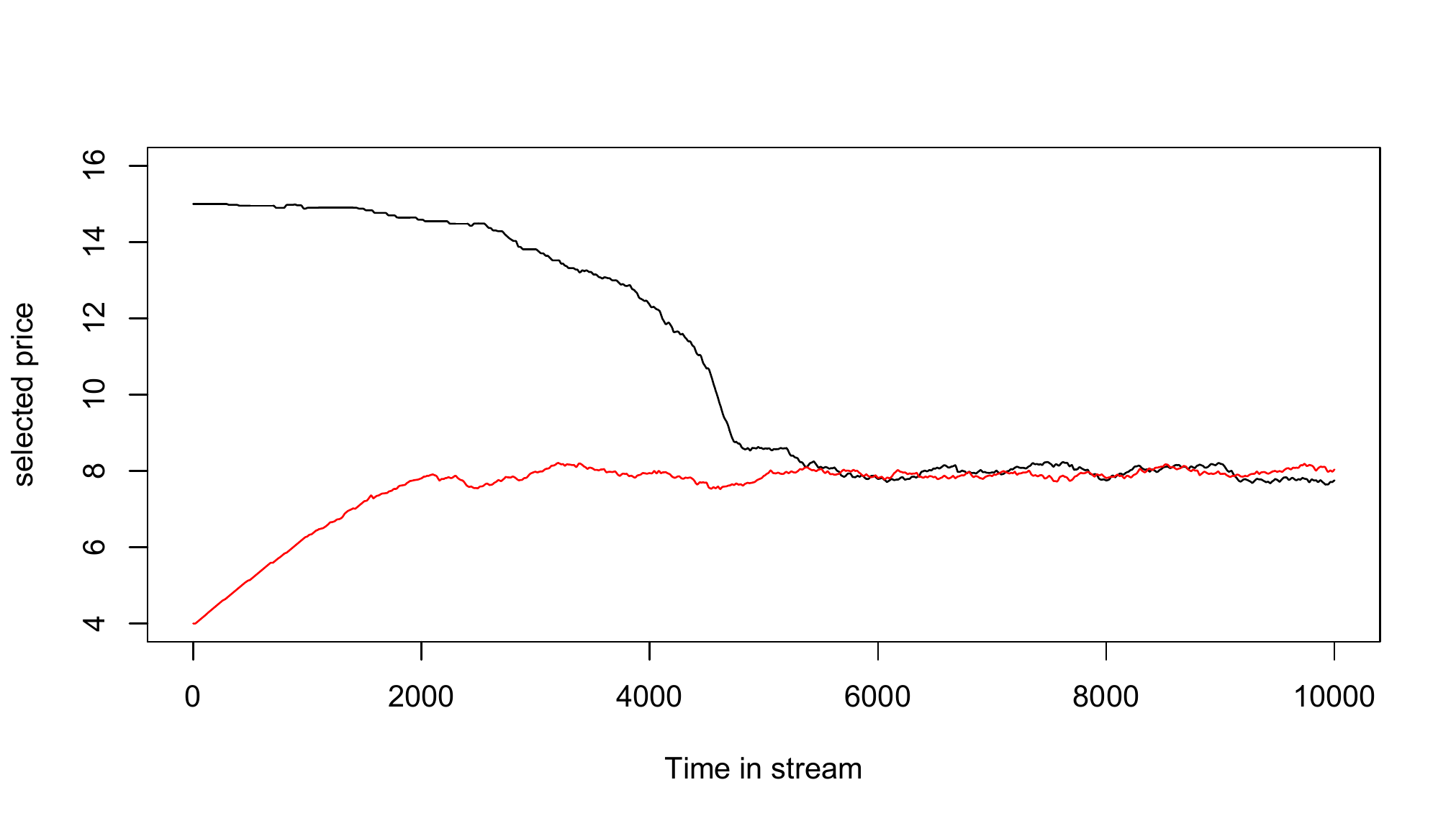}
      \caption{Use of LiF to find the revenue maximizing sales-price for a firm: example of a setup in which the observed $y \in \{0,1\}$.}
          \label{fig:price}
\end{figure}

Figure \ref{fig:price} shows the performance of LiF-\rom{2} for two different starting values, $x_0 = 4$ and $x_0 =15$, using the same set of tuning parameters as those used in Study 3 ($t=10^4=\mathcal{T}$, $A=1$, $T=100$, $\gamma=.1$). The only change in the algorithm compared to the earlier simulations is that $r_t = y_t x_i$ is integrated (summed) over instead of using the observed $y_t$ directly. Also in this case, LiF finds $x_{max}$ fairly quickly (in under 6000 iterations). 

It has to be noted that too high starting values, and thereby a very low $Pr(y = 1 | x_0)$ might lead to a failure to find $x_{max}$ since LiF then get's stuck in a local ``maximum'': for very high values of $x$ the revenue $r$ will always be $0$.

\section{Simulation study 5: Empirical Regret}

The previous studies show that LiF is effective in finding the value of $x_{max}$. However, the oscillation that is introduced clearly introduces search costs into the procedure: LiF continuously runs experiments with a certain amplitude in its variation in $x$ to find $x_{max}$. In the previous simulations these search costs have not been considered, and hence while these simulations demonstrate that LiF finds the value of $x_{max}$, the previous simulation studies are uninformative regarding the costs of the procedure. To address this problem we run another simulation study in which we monitor the empirical \emph{regret} 
\begin{align}
\mathcal{R}(t) = \sum_{i=1}^t (f(x_{max}) - (f(x_t))
\end{align} 
of the procedure. Thus, we compare over time in the data stream how much ``is lost'' when using LiF as compared to always selecting the exact right value of $x$ that maximizes the outcome if the data generating process would have been known. We use the exact setup as used in Simulation Study 4 (exact same data generating model and tuning parameter settings), but we increase $\mathcal{T}$ to $10^5$. Also, because of the noise and our interest in LiF as a general procedure, not merely in one specific attempt, we replicate the simulation $M=100$ times.

To give insight in the performance of LiF when examining the regret of the procedure, we contrast the use of LiF not only to selecting the optimal value, but also to two other sequential experimentation scheme's:
\begin{itemize}
\item \emph{$\epsilon$-first:} in this approach we run a limited time (up to $n=1000$) experiment in which we randomly sample values of $x$ uniformly between 0 and 20. Subsequently, we fit a simple logistic regression modeling $Pr(y=1|x) = \mathcal{L}(\beta_0 + \beta_1 x)$ where $\mathcal{L}()$ denotes the logit link (see also Equation \ref{eq:logit}), and determine $x_{\epsilon} =  \arg\max_{x}  \mathcal{L}(\beta_0 + \beta_1 x) x$. The remaining $\mathcal{T} - n$ observations in the stream are allocated to $x_{\epsilon}$.
\item \emph{Bootstrap Thompson Sampling (BTS)}: In this sequential experimentation scheme we again fit a simple logistic regression to estimate $Pr(y=1|x)$. We use Stochastic Gradient Descent to update the parameters of the model at each time point during the data stream. Furthermore, we maintain $J=100$ models each using an online half-sampling bootstrap to perform bootstrap Thompson sampling \citep[See for details of this sequential allocation scheme][]{Kaptein}. This gives $J$ different estimates of the model parameters ($\{\beta_0^j, \beta_1^j\}$). We then randomly uniformly select $j'$ out of $j=1, \dots, j=J$ and select treatment $x_{bts} =  \arg\max_{x} \mathcal{L}(\beta_0^{j'} + \beta_1^{j'} x) x$. This bootstrapped sampling scheme quantifies the uncertainty in the model estimates and uses this directly to balance exploration (querying new values for $x$ to learn more about the data-generating model), and exploitation (selecting the value of $x$ which one believes leads to the highest outcomes). 
\end{itemize}
Note that we choose random starting points of the parameter values for BTS that are relatively close to the true values, and that the functional form of the model that is used is the same as the true data generating model. Hence, this latter condition is expected to do very well on the current problem since it implements a lot of knowledge regarding the data-generating function that is not accessible to LiF.

\begin{figure}[!ht]
  \centering
    \includegraphics[width=0.75\textwidth]{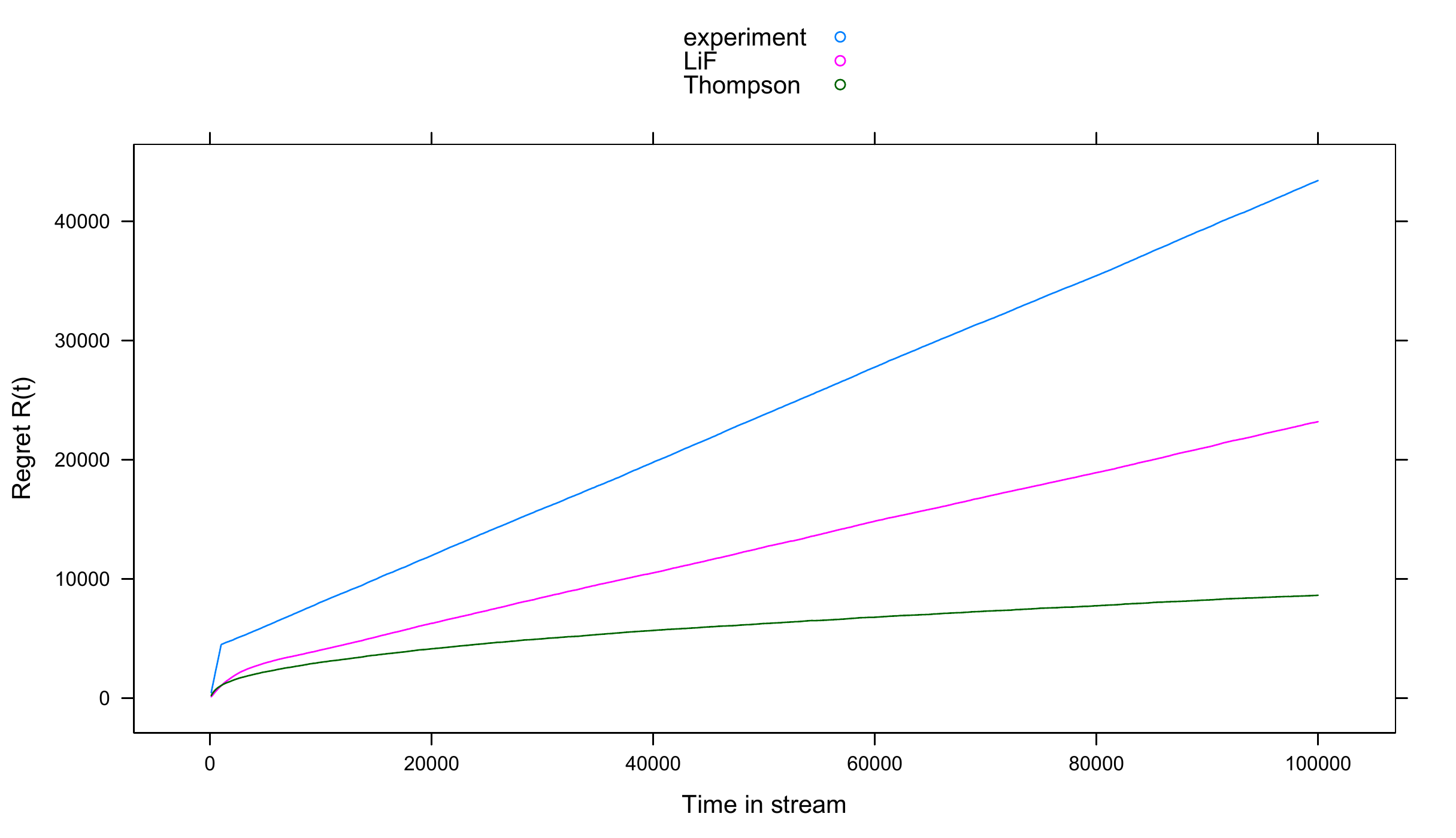}
      \caption{Overview of the (mean) empirical regret of three possible sequential allocation schemes.}
          \label{fig:regret}
\end{figure}

Figure \ref{fig:regret} shows the performance of LiF-\rom{2} -- in terms of average regret -- compared to the $\epsilon$-first and BTS. It is clear that the $\epsilon$-first does not perform very well: logically, during the experimentation stage $t=1, \dots, t=n$ this method incurs a large regret. However, since the probability that the true $x_{max}$ is found exactly in the experiment is smaller then $1$, also after the experiment period (expected) linear regret is incurred. BTS performs much better in the long run: the regret is not linear but rather seems to be $\mathcal{O}(\sqrt(t))$, which is the proven minimal regret bound known for this problem \citep{NIPS2011_4475}.

Early on LiF performs very well on this problem; LiF is very efficient in finding $x_{max}$. It is even more efficient than BTS for small $t$, despite the fact that in the current setup BTS is heavily favored by using the correct form of the data generating model, something which is in practice very unlikely. However, in the long run the regret of LiF is lineair in $t$. This latter fact is easily explained: due to the continuous oscillations of $x$ by adding $A \cos \omega t$ LiF keeps exploring the space and thus keeps incurring additional costs. Even if $x_{max}$ has been found, these search costs are linear with $t$.

This simulation suggests that, in the bandit feedback case, LiF can be improved by gradually decreasing the amplitude of the oscillation: if $A$ can be decreased as a function of (e.g.,) the approximated gradient as well as the current time in the stream, the exploration behavior of LiF can be systematically decreased over time in the stream. However, this would make LiF less sensitive to concept drift, which might in practice be infeasible. Hence, we currently regard the linear regret incurred by LiF as exploration costs necessary to ensure its robustness in a changing environment.

\section{Discussion and Future work}

In this paper we presented Lock in Feedback as a method to find $\arg\max_{x} f(x)$ through sequential experiments. The method is appealing since it a) does not require the functional form of $f(x)$ to be known to derive its maximum, b) performs well in situations in which measurements are obtained with large noise, and c) allows following the maximum of a function even if that function changes over time. We have presented the basic mathematical arguments behind LiF, demonstrating how known (or imposed) oscillations in $x$ can be used to determine the derivative(s) of $f(x)$ which can subsequently be used to find $\arg\max_{x} f(x)$. Next, we detailed two possible implementations of LiF and examined their performance for a variety of tuning parameter settings. We then showed that a streaming version of LiF is robust both to noise as well as concept drift.

We believe LiF can be of use in many sequential experimentation problems in which the independent variable is continuous; in the introduction we discussed pricing, medication dosing, and the selection of items by their difficulty as possible examples. LiF is extremely easy to implement, and very robust to noise and concept drift. We thus hope that LiF can be a valuable tool in treatment optimization in sequential experiments.

However, the current expose of LiF also introduces a number of questions. For example, the ability to use LiF for problems of higher dimensions, e.g., where $y = f(\vec{x})$ is a function of multiple variables, has not been explored here even though this extension relatively is easily made. Also, the suggested decrease of the amplitude in the bandit setting (Simulation Study 5) needs further scrutiny and begs for an analytical treatment of the use of LiF in stochastic optimization with bandit feedback \citep[see, e.g.,][]{NIPS2011_4475}. Finally, the currently proposed version of LiF allows one to find local maxima (or minima), but convergence to a global maximum is not guaranteed. Throughout this paper, we have been considering unimodal functions, which might, in practical applications, be a too stringent assumption.

In this paper we have demonstrated the use of LiF only in cases where $x$ is scalar. However, when $x$ is a vector a very similar approach can be used to find the maximum of the function $f(\vec{x})$ in more than one dimension. In the two dimensional case LiF can be extended by oscillating both elements of $x$ at different frequencies:
\begin{align*}
x_{1,t} = x_{1,0} + A_1 \cos{\omega_1 t} \\
x_{2,t} = x_{2,0} + A_2 \cos{\omega_2 t}
\end{align*}
After oscillating both elements of $x$ we observe $y_t = f(x_{1,t}, x_{2,t})$ and we can obtain information regarding the gradient by separately computing:
\begin{align*}
y_{1,\omega} = y_t \cos{\omega_1 t} \\
y_{2,\omega} = y_t \cos{\omega_2 t}
\end{align*}
This simple extension allows for the use of LiF in higher dimensions. However, besides the fact that $\omega_1$ and $\omega_2$ should not be multiples of each other, the effects of the tuning parameters and the performance of this higher dimensional version of LiF need to be further examined.

Our proposed LiF algorithm, similar to many other procedures for function maximization, is prone to uncovering local maxima instead of global maxima. A logical solution to this problem would be to consider multiple starting points $\vec{x}_0$ which are oscillated independently (possibly alternating within a data stream). Effectively this would allow the experimenter to find multiple maxima. By evaluating the value of $y$ one could decide on the best possible solution, or, one could pool the results of multiple alternating threats to update each of them. Such approaches, and their robustness to the existence of local maxima, needs further scrutiny.

\section*{References}
\bibliographystyle{apa}
\bibliography{library}
\clearpage

\section*{Algorithm for finding the exact maximum of a parabola using the second order approximation.}
\label{app:exact}
Let's suppose that the curve $y=f(x)$ is a parabola:
\begin{align*}
y=-\alpha(x-x_0)^2+\gamma
\end{align*}
Clearly, $f(x)$ has a maximum for $x=x_0$. Furthermore, the second derivative is always equal to $-2\alpha$, regardless the value of $x$. Interestingly, the value of $\alpha$ can be easily extracted from the data accumulated during the lock-in procedure. For this purpose, $y(t)$ has to be multiplied by $\cos\left( 2\omega t \right)$. Following the steps illustrated in eq. \ref{complete}, eq. \ref{rewrite}, and eq. \ref{rewrite2}, one obtains:
\begin{align*}
y_{2\omega} & = \frac{TA^2}{8}\left(\left.  \frac{\partial^2 f}{\partial x^2}\right|_{x=x_0}\right) \\
		& + \int_{0}^{T} \epsilon \cos\left( 2\omega t \right)
\end{align*}
which allows us to calculate $\alpha$ as:
\begin{align*}
\alpha=\frac{4 y_{2\omega}}{TA^2}
\end{align*}
\clearpage

\end{document}